\documentclass[sigconf]{acmart}
\usepackage{booktabs}
\usepackage{multirow}
\usepackage{enumitem}
\usepackage{siunitx}
\usepackage[ruled,linesnumbered]{algorithm2e}
\usepackage{subcaption}
\usepackage{hyperref}
\hypersetup{
    colorlinks = true,
    citecolor  = blue,
    linkcolor  = blue,
    urlcolor   = blue,
}
\usepackage{caption}

\def\sectionautorefname~#1\null{Section~#1\null}

\copyrightyear{2027}
\acmYear{2027}
\setcopyright{cc}
\setcctype{by}
\acmConference[ASPDAC '27]{32nd Asia and South Pacific Design Automation Conference}{January 25--28, 2027}{Tokyo, Japan}
\acmBooktitle{32nd Asia and South Pacific Design Automation Conference (ASPDAC '27), January 25--28, 2027, Tokyo, Japan}
\acmDOI{10.1145/3840404.3848324}
\acmISBN{979-8-4007-2923-2/2027/01}
\AtBeginDocument{%
  }

\hypersetup{
    colorlinks=true,
    urlcolor=blue,
    linkcolor=black,
    citecolor=black
}
\begin{document}
\raggedbottom

\title{Topology as a Design Variable: Joint Architecture-Sizing Optimization for SRAM}







\author{Can Wang}
\affiliation{\institution{Nanjing University of Sci.\&Tech.}\city{Nanjing}\country{China}}
\email{wangcan\_01@njust.edu.cn}

\author{Junhao Ma$^+$}
\affiliation{
  \institution{Institute of Microelectronics of the Chinese Academy of Sciences}
  \city{Beijing}
  \country{China}
}

\email{majunhao26@mails.ucas.ac.cn}

\author{Yikai Wang$^+$}
\affiliation{
  \institution{Institute of Microelectronics of the Chinese Academy of Sciences}
  \city{Beijing}
  \country{China}
}

\email{wangyikai26@mails.ucas.ac.cn}

\author{Shan Shen$^*$}
\affiliation{\institution{Nanjing University of Sci.\&Tech.}\city{Nanjing}\country{China}}
\email{shanshen@njust.edu.cn}

\author{Wei W. Xing$^*$} 
\affiliation{\institution{University of Sheffield}\city{Sheffield}\country{United Kingdom}}
\email{w.xing@sheffield.ac.uk}

\thanks{
This work was supported by the Fundamental Research Funds for the Central Universities (Grant No. 30925010605, 30924012004), Jiangsu Provincial Major Science and Technology Project (Grant No. BG2025012). $^+$ This work was conducted when Junhao Ma and Yikai Wang were with Nanjing University of Science and Technology, Nanjing, China.
$^*$Corresponding authors: S. Shen, W. Xing.
}


\begin{abstract}
SRAM's power-performance-area-stability trade-off depends on bitcell topology, a discrete architectural choice that interacts non-separably with continuous transistor sizing. However, no open-source SRAM tool treats topology as an optimizable variable. OpenRAM generates SRAM layouts but performs no optimization, while OpenYield supports high-fidelity evaluation but restricts optimization to scalar sizing of a fixed 6T cell. OpenYieldV2 closes this gap through three integrated contributions. First, a fully parameterized compiler generates complete 6T and 10T SRAM macros with peripheral circuits and replica-bitline self-timing, enabling topology comparison under identical simulation conditions. Second, a few-shot TabPFN surrogate learns topology-dependent PPA responses with limited full-macro SPICE samples. Third, a coarse-to-refined multi-objective optimizer combines global exploration of discrete architectural choices with local refinement of continuous transistor sizes. Experiments across varying design conditions reveal a topology decision boundary that shifts with operating constraints rather than following a fixed design rule. Jointly optimizing across this boundary improves hypervolume by up to 17.3\% over the strongest single-topology baseline. \href{https://github.com/ShenShan123/OpenYield.git}{OpenYieldV2} is open-source to support reproducible SRAM design research.
\end{abstract}

\maketitle

\section{Introduction}
SRAM occupies up to 70\% of die area in modern AI accelerators and remains the dominant on-chip memory technology across CPU caches, SoCs, and ASICs~\cite{OpenYield}. Designing it well requires solving a genuinely mixed problem: a discrete choice of bitcell topology and per-transistor threshold-voltage assignment that interacts non-separably with continuous transistor sizing.
Yet the SRAM optimization literature, open-source and academic alike, has grown quite sophisticated at answering a narrower question than this. 
Given a topology, fixed in advance, what is the best transistor sizing? 
The more consequential decision---which topology to use---is treated as a fixed input rather than something the optimizer is allowed to explore. 
This is not a minor omission, since sizing and topology are not separable in SRAM: a pull-down transistor's optimal width under a 6T cell has little bearing on its optimal width under a Schmitt-trigger 10T cell, because the trip-point dynamics the sizing must satisfy are different. 
Optimizing one while fixing the other explores only a slice of the design space.
Making topology a decision variable, rather than an assumption fixed in the netlist, appears to be a simple extension of existing sizing optimization. However, the difficulty compounds in three stages, each exposed only after the previous one is addressed.

The first obstacle is trust. Comparing two topologies is meaningful only when the evaluation reflects the actual circuits. Simplified models that omit interconnect parasitics, inter-cell leakage, or peripheral circuitry underestimate read delay by up to $3.7\times$ on a 64-row array and write delay by $8.45\times$ when full peripheral circuits are included~\cite{OpenYield}. A topology comparison based on such models answers a different question. OpenRAM~\cite{guthaus2016openram} generates full layouts of 6T and 10T topologies but performs no optimization, leaving topology selection to the user. OpenYield~\cite{OpenYield} closes the fidelity gap with distributed parasitics and full-peripheral simulation, but its optimization remains restricted to scalar sizing on a fixed 6T cell. Thus, high fidelity is achieved only after the topology has already been selected.

Suppose fidelity is addressed, as OpenYield's infrastructure largely does. A second obstacle, efficient surrogate modeling, then emerges. Once topology is unfrozen, the optimizer must learn multiple topology-dependent sizing surfaces, while every training sample still requires SPICE simulation. OpenOpt~\cite{wang2026openopt} reduces per-evaluation cost through an equivalent-circuit model, rather than a learned surrogate, and therefore does not address cross-topology surrogate learning. Existing surrogate-assisted SRAM methods remain topology-specific: Lee et al.~\cite{lee2023sram} use Gaussian-process Bayesian optimization for power and access-time sizing, and Kim et al.~\cite{kim2025sram} extend it to multiple instances and PVT corners. Related mixed-variable methods either embed discrete device choices into a continuous space for Gaussian-process optimization~\cite{devrep2022} or directly search continuous and categorical variables without a surrogate~\cite{catcma2024}. Neither provides a tuning-free surrogate for topology-dependent PPA evaluation of complete SRAM macros, including their sense amplifiers, decoders, and replica-bitline timing.

A third obstacle is the optimization strategy itself. SRAM design is inherently multi-objective because delay, power, area, and noise margin conflict, whereas a scalar objective can conceal useful Pareto trade-offs. Prior optimizers address only parts of this problem. OpenYield~\cite{OpenYield} performs single-objective transistor sizing on a fixed 6T cell, while OpenOpt~\cite{wang2026openopt} supports both single-objective algorithms and NSGA-II/MOEA/D but keeps the bitcell topology fixed. Conventional evolutionary algorithms can explore discrete choices and multiple objectives, yet converge slowly on continuous sizing variables because crossover and mutation do not exploit local geometry. Conversely, continuous analog-sizing methods such as MACE~\cite{lyu2018mace} cannot directly handle discrete topology choices. Joint topology--sizing optimization therefore requires multi-objective global exploration of discrete architectures followed by dedicated local refinement of continuous transistor sizes.

No existing open-source tool has addressed all three obstacles simultaneously, and solving them individually only reveals the next challenge. This paper presents \textbf{OpenYieldV2}, which addresses them together with the following contributions:

\begin{itemize}[topsep=0pt, partopsep=0pt]
    \item \textbf{A dual-topology, fully parameterized SRAM compiler.} Complete peripheral circuits, including replica-bitline self-timing, alongside native 6T and 10T Schmitt-trigger bitcells, producing simulation-faithful macros for either topology under identical high-fidelity conditions.

    \item \textbf{A few-shot surrogate for multi-topology search.} TabPFN~\cite{TabPFN} requires no dataset-specific tuning and 
    achieves a mean Kendall rank correlation above 0.94 with 300 SPICE samples, whereas all competing surrogates remain below 0.80 even with 1{,}000 samples.

    \item \textbf{A coarse-to-refined optimizer suited to mixed variables.} Discrete topology and threshold-voltage choices are resolved through surrogate-guided evolutionary search; continuous sizing is then refined within each architecture via residual-adaptive gradient descent.

    \item \textbf{The first characterization of the SRAM topology preference boundary.} Joint topology optimization reveals that topology selection cannot be reduced to a fixed design-time rule: 6T and 10T jointly shape the Pareto frontier under most design conditions, while topology-dominant regions emerge only under stringent requirements and shift with changing conditions.
    
\end{itemize}

\section{Background}
\label{sec:background}

\subsection{SRAM Compilers and Surrogate Modeling}

Commercial memory compilers, such as TSMC's 16nm FinFET generator~\cite{TSMC2014} and Intel's architecture-flexible tool~\cite{Intel2001}, size and lay out SRAM macros with production-grade accuracy, but the topology and sizing strategy they use are fixed inside a proprietary flow that never exposes them to optimization. Open-source alternatives remove the opacity but not the freeze. OpenRAM~\cite{guthaus2016openram} generates full macro layouts across several bitcell topologies, including 6T and 10T, yet performs no optimization at all: topology and every transistor dimension are supplied by the user and never revisited. OpenYield~\cite{OpenYield} closes the fidelity gap that made such comparisons untrustworthy in the first place, showing that models omitting distributed parasitics and peripheral-circuit variation understate read delay by up to $6.87\times$ and write delay by $8.45\times$, but its optimization platform operates on a single, pre-selected 6T cell. The topology decision that fidelity was built to inform is never exposed to the optimizer that follows it, and no open-source tool, at present, evaluates competing topologies under the same high-fidelity conditions it uses to size any one of them.

Unfreezing topology expands the search space, while each training sample still requires a full SPICE run. Gaussian-process surrogates reduce the simulation budget by learning from a small set of samples~\cite{lee2023sram,kim2025sram}, but updating the model as samples accumulate incurs kernel fitting costs. In our prior work on multi-corner SRAM yield analysis, TabPFN used learned priors to adapt to new circuit data through in-context inference without per-circuit hyperparameter tuning~\cite{xing2026breaking}. Here, we use TabPFN~\cite{TabPFN} as a surrogate for topology-dependent SRAM PPA: it treats simulated designs as context and predicts candidate performance without fitting a new kernel after each batch. This makes surrogate updates practical as the search spans multiple bitcell topologies.

\subsection{Joint Topology-Sizing Optimization}
Even with an affordable surrogate, an optimizer still has to search a space made of two different kinds of variables. Evolutionary algorithms handle the discrete part comfortably but converge slowly on the continuous sizing surface within any one architecture, since crossover and mutation do not exploit the differentiable structure sizing actually has. Methods built for analog synthesis have the opposite weakness: MACE~\cite{lyu2018mace} and related multi-objective Bayesian approaches search continuous sizing efficiently but are not built to accommodate a discrete topology choice at all. A separate line of analog IC work does search topology and sizing jointly, typically by embedding discrete device choices into a continuous latent space that a Gaussian process can then explore~\cite{devrep2022,catcma2024}, but these methods evaluate their objective from a single device or cell, whereas an SRAM macro's delay, power, and noise margin depend on the sense amplifiers, decoders, and replica-bitline timing that surround it, none of which a single-device embedding was built to capture.

Accordingly, joint topology and sizing optimization can be formulated as a mixed-integer multi-objective problem over a joint space $\mathcal{X} = \mathcal{X}_d \times \mathcal{X}_c$, where $\mathbf{x}_d \in \mathcal{X}_d$ collects the discrete architectural choices (bitcell topology, per-transistor threshold-voltage assignment, array organization) and $\mathbf{x}_c \in \mathcal{X}_c$ collects the continuous sizing parameters:
\begin{equation}
\begin{aligned}
\min_{\mathbf{x} \in \mathcal{X}} \quad
& \mathbf{F}(\mathbf{x}) = \bigl(f_{\text{delay}}(\mathbf{x}),\,
  f_{\text{power}}(\mathbf{x}),\, f_{\text{area}}(\mathbf{x}),\,
  {-f_{\text{snm}}(\mathbf{x})}\bigr)^T, \\
\text{s.t.} \quad
& g_k(\mathbf{x}) \le 0, \quad k = 1, \dots, K,
\end{aligned}
\label{eq:formulation}
\end{equation}
with $g_k$ encoding constraints such as delay upper bounds or SNM lower bounds under specific PVT and Monte Carlo conditions. Section~\ref{sec:framework} describes how OpenYieldV2 solves this problem: a compiler that makes $\mathbf{x}_d$ and $\mathbf{x}_c$ simulatable under the same high-fidelity conditions, a surrogate that keeps evaluating $\mathbf{F}$ affordable as $\mathcal{X}$ grows, and an optimizer that searches $\mathcal{X}_d$ and $\mathcal{X}_c$ with the method suited to each.



\section{The OpenYieldV2 Framework}
\label{sec:framework}
\begin{figure}[t]
    \centering
    \setlength{\abovecaptionskip}{1pt}
    \includegraphics[width=1.0\linewidth]{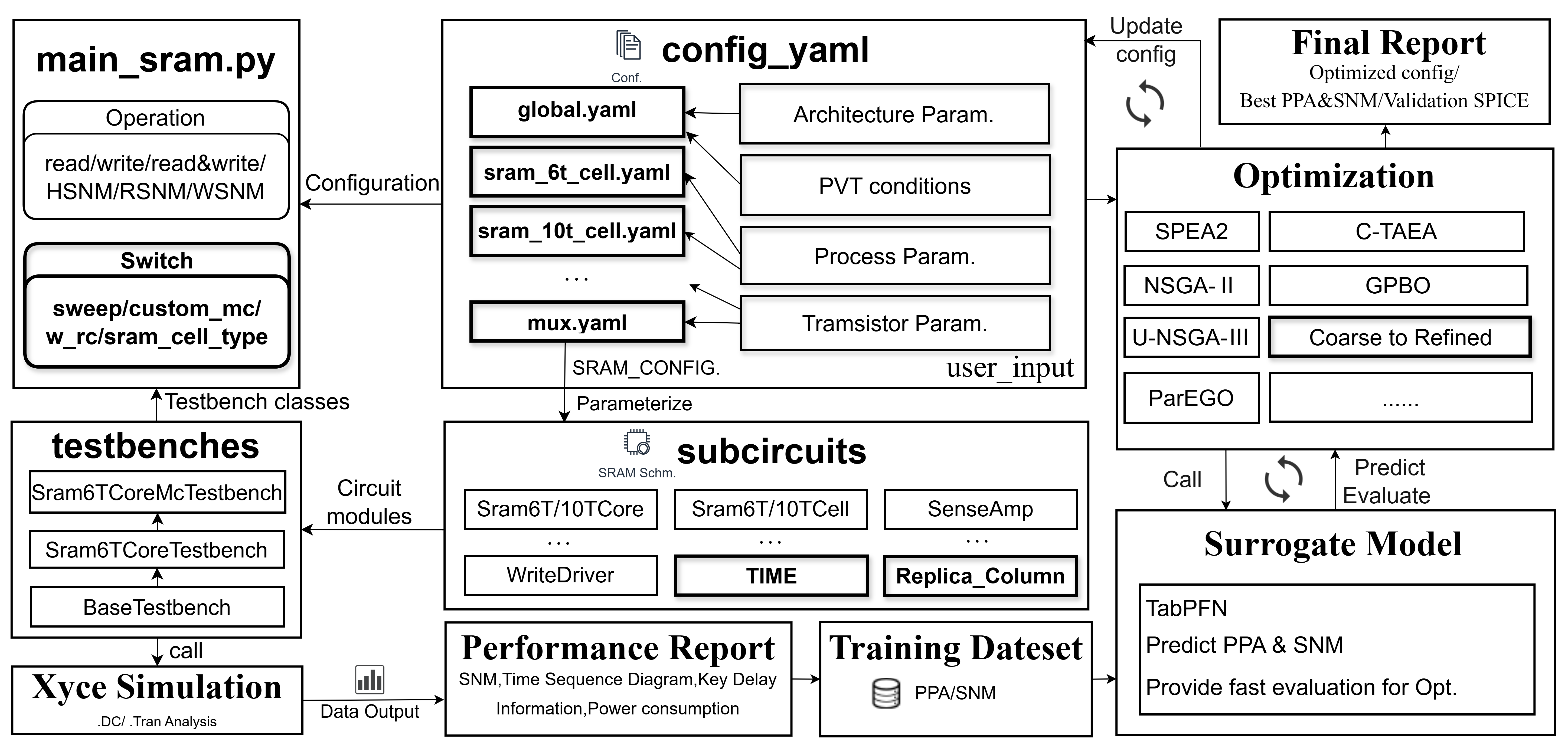}
    \caption{The OpenYieldV2 pipeline. The main entry point, \texttt{main\_sram.py}, reads YAML configurations and invokes testbenches, which instantiate SRAM circuits from \texttt{subcircuits} and run Xyce simulations. Simulation results support surrogate modeling and optimization.}
    \label{fig:Code Architecture}
\end{figure}

The full pipeline is shown in \autoref{fig:Code Architecture}.
OpenYieldV2 takes a single configuration, array dimensions, PDK, PVT corners, process variation statistics, and sizing bounds, and turns it into SPICE netlists, PPA metrics, and optimized designs through a closed loop. \texttt{main\_sram.py} coordinates netlist construction via PySpice~\cite{PySpice}, stimulus generation, and parallelized simulation through Xyce~\cite{keiter2013xyce}. What distinguishes this loop from a conventional sizing flow is what sits inside it: the population being evolved carries topology and threshold-voltage assignment alongside every continuous sizing parameter, so a candidate design is never evaluated against a topology fixed in advance. SPICE outputs feed a TabPFN surrogate that keeps this joint search affordable, and the surrogate in turn drives a coarse-to-refined optimizer that treats the discrete and continuous parts of $\mathbf{x}$ with the method suited to each.

\subsection{Dual-Topology Bitcell Generation}

The generator supports two bitcell topologies with distinct operating regimes. The conventional 6T cell stores data in a cross-coupled inverter pair accessed through two pass transistors; its compact layout suits area-sensitive designs, but the pull-down transistor must be sized to overpower the access transistor during a read, which limits how aggressively sizing can be pushed at reduced supply voltage. The 10T cell replaces the storage inverters with Schmitt-trigger circuits~\cite{10T}, whose topology is depicted in \autoref{fig:sram_circuit_organization}. It introduces a feedback-dependent switching threshold that shifts the cell's trip point away from the disturb direction during a read and tolerates larger threshold-voltage shifts before failing. The cost is a longer read delay from the added series resistance in the Schmitt-trigger pull-down path, quantified in Section~\ref{sec:experiments}. Both topologies, along with per-transistor threshold-voltage assignment (LVT, RVT, HVT), are exposed as discrete inputs to the same netlist generator, so switching between them changes only the array configuration passed to \texttt{main\_sram.py}, not the simulation or optimization code that consumes the output. This is what makes topology simulatable on equal footing with sizing: the generator produces both circuits under identical PVT, Monte Carlo, and peripheral conditions, so any difference the optimizer later observes between them reflects the circuits themselves rather than an inconsistency in how they were built.

\subsection{High-Fidelity Peripheral Circuits}

\begin{figure}[t]
    \centering   \includegraphics[width=1.0\linewidth]{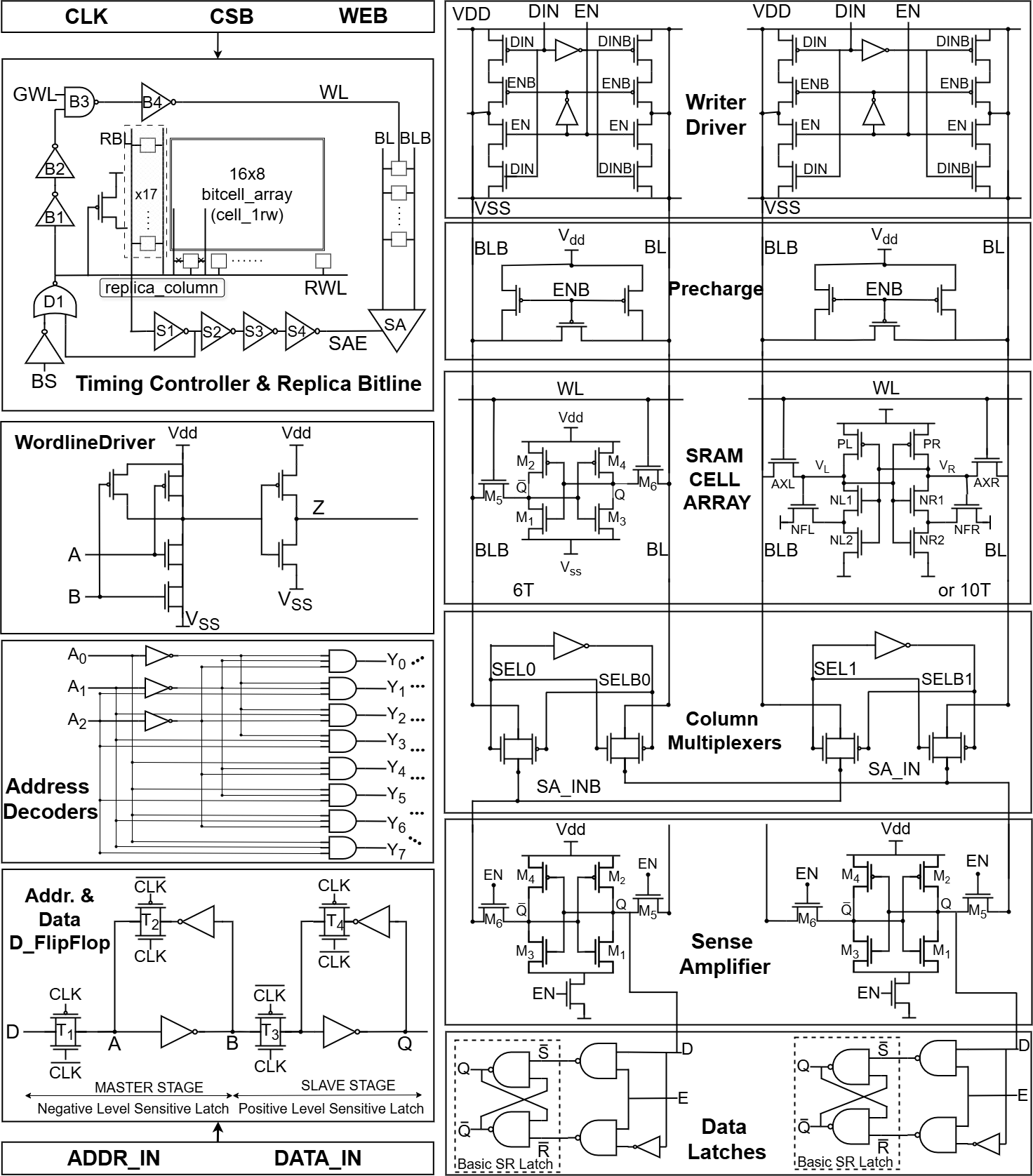}
    \caption{SRAM macro organization in OpenYieldV2. The bitcell array is surrounded by complete peripheral circuits, generated identically regardless of which bitcell topology is selected.} \label{fig:sram_circuit_organization}
\end{figure}

A bitcell array evaluated in isolation does not reflect the delay, power, or noise margin of a working macro, and OpenYieldV2 surrounds the array accordingly. As shown in \autoref{fig:sram_circuit_organization}, the generator instantiates modular 3-to-8 address decoders cascaded to cover arbitrary address spaces, precharge circuits for bitline equalization, 2:1 column multiplexers, voltage-latched differential sense amplifiers, tri-state write drivers, wordline drivers, transmission-gate flip-flops, and D latches, with every driving transistor sized automatically to the array dimensions. Timing is generated through a replica-bitline (RBL) controller~\cite{replica} rather than a fixed delay chain: the replica bitline carries a scaled fraction of the full bitline parasitic and is discharged by a replica cell under the same conditions as the accessed cell, so the controller's three internal signals, precharge enable, sense amplifier enable, and write enable, track actual bitline discharge across PVT corners instead of a margin calibrated to one corner.  \autoref{sec:background} already established why this completeness matters; what it buys here is a common, trustworthy evaluation surface on which 6T and 10T can be compared, which the next two sections build on.

\subsection{Few-Shot Surrogate Modeling}
Full-macro SPICE evaluation is costly, motivating a surrogate for
joint topology--sizing search. We use TabPFN~\cite{TabPFN}, a
pretrained tabular model that predicts from labeled examples
provided as context, without updating its pretrained weights in
our implementation. Let
$\mathcal{D}=\{(\mathbf{x}_i,\mathbf{y}_i)\}_{i=1}^{N}$
denote the SPICE-labeled context set. For each performance metric
$j$, we construct
$\mathcal{D}_j=\{(\mathbf{x}_i,y_{ij})\}_{i=1}^{N}$
and obtain a separate prediction using the same pretrained
backbone. Our implementation converts the model's output-bin
probabilities into a point prediction:
\begin{equation}
\widehat{y}_j(\mathbf{x})
=
\bar{y}_j+s_j
\sum_{b=1}^{B_j}
p_{j,b}(\mathbf{x}\mid\mathcal{D}_j)\,c_{j,b},
\end{equation}
where $p_{j,b}$ is the predicted probability of bin $b$,
$c_{j,b}$ is its center in standardized target space, and
$\bar{y}_j$ and $s_j$ are the target normalization statistics.
This avoids dataset-specific optimization of the pretrained
weights or Gaussian-process kernel hyperparameters.

The input representation includes topology information and
topology-specific design variables, allowing 6T and 10T samples
to be pooled within each metric's context. In the optimization
experiments, this labeled context is established before the
search and remains fixed throughout each run. The optimizer
queries the surrogate to evaluate candidate designs and
approximate Pareto trade-offs; these queries do not add new
SPICE-labeled samples to $\mathcal{D}$, and uncertainty-based
acquisition is not used. Thus, the implemented workflow is
fixed-context surrogate-assisted optimization rather than an
online active-learning loop. Pooling samples from both supported
topologies enables joint search without separately tuning a
surrogate for each topology.

\subsection{Joint Topology-Sizing Optimization}
Discrete and continuous variables require different search strategies.
Let $\mathbf{x}=(\mathbf{d},\mathbf{c})$, where $\mathbf{d}$ collects
discrete choices, including bitcell topology, threshold-voltage
assignment, and array organization, while $\mathbf{c}$ contains
continuous transistor-sizing parameters.
The surrogate objective vector $\hat{\mathbf{F}}$ contains $M$
objectives, all expressed in minimization form.
Evolutionary search explores these mixed variables without requiring
gradients, but does not directly exploit the surrogate's
differentiability with respect to sizing.
OpenYieldV2 therefore combines cross-topology coarse search with
gradient-based refinement within each selected discrete configuration.

The coarse stage jointly explores $\mathbf{d}$ and $\mathbf{c}$
through crossover, mutation, and multi-objective selection.
The feasible candidates evaluated during this search yield a
surrogate-predicted non-dominated set $\mathcal{P}_{\mathrm{ND}}$,
representing different design trade-offs.
We select up to $K$ representative seeds distributed across the normalized
objective space to avoid refining near-duplicate solutions.
Each seed $\mathbf{x}_k^\star=(\mathbf{d}_k^\star,\mathbf{c}_k^\star)$
retains the discrete configuration identified by the coarse search.
Here, $\star$ denotes a selected coarse seed rather than a guaranteed
optimum.
These seeds provide both the initial designs and the objective
references for subsequent refinement.
The coarse-search and seed-selection procedure is summarized
in Algorithm~\ref{alg:coarse_search}.

\begin{algorithm}[t]
\caption{Coarse Search for Representative Seeds}
\label{alg:coarse_search}
\KwIn{Surrogate $\hat{\mathbf{F}}$, population size $N$,
surrogate-evaluation budget $B\geq N$, seed limit $K$}
\KwOut{Seed set $\mathcal{P}_{\mathrm{coarse}}$}

$\mathcal{P}\leftarrow\textsc{MixedInitialize}(N)$\;
\textsc{EvaluateAndStore}$(\mathcal{P},\hat{\mathbf{F}})$\;
$\mathcal{H}\leftarrow\mathcal{P}$,\quad $b\leftarrow N$\;

\While{$b<B$}{
    $\mathcal{Q}\leftarrow
    \textsc{MixedVariableMating}(\mathcal{P})$\tcp*{crossover,
    mutation, and duplicate handling}
    $\mathcal{Q}\leftarrow
    \textsc{LimitBatch}(\mathcal{Q},B-b)$\;
    \textsc{EvaluateAndStore}$(\mathcal{Q},\hat{\mathbf{F}})$\;
    $\mathcal{H}\leftarrow\mathcal{H}\cup\mathcal{Q}$\;
    $b\leftarrow b+|\mathcal{Q}|$\;
    $\mathcal{P}\leftarrow
    \textsc{SMSEMOASurvival}(\mathcal{P}\cup\mathcal{Q},N)$\;
}

$\mathcal{P}_{\mathrm{ND}}\leftarrow
\textsc{UniqueNonDominated}(\textsc{Feasible}(\mathcal{H}))$\;
$\mathcal{P}_{\mathrm{coarse}}\leftarrow\varnothing$\;
\If{$\mathcal{P}_{\mathrm{ND}}=\varnothing$}{
    \Return{$\mathcal{P}_{\mathrm{coarse}}$}\;
}

$\{\mathbf{z}(\mathbf{x})\}\leftarrow
\textsc{NormalizeForDistance}(\mathcal{P}_{\mathrm{ND}})$\;
$K'\leftarrow\min(K,|\mathcal{P}_{\mathrm{ND}}|)$\;
$\mathbf{x}_1^\star\leftarrow
\underset{\mathbf{x}\in\mathcal{P}_{\mathrm{ND}}}{\arg\min}\,
\frac{1}{M}\sum_{i=1}^{M}z_i(\mathbf{x})$\;
$\mathcal{P}_{\mathrm{coarse}}\leftarrow
\{\mathbf{x}_1^\star\}$\;

\While{$|\mathcal{P}_{\mathrm{coarse}}|<K'$}{
    $\mathbf{x}^\star\leftarrow
    \underset{\mathbf{x}\in
    \mathcal{P}_{\mathrm{ND}}\setminus\mathcal{P}_{\mathrm{coarse}}}
    {\arg\max}\,
    \min_{\mathbf{u}\in\mathcal{P}_{\mathrm{coarse}}}
    \|\mathbf{z}(\mathbf{x})-\mathbf{z}(\mathbf{u})\|_2$\;
    $\mathcal{P}_{\mathrm{coarse}}\leftarrow
    \mathcal{P}_{\mathrm{coarse}}\cup\{\mathbf{x}^\star\}$\;
}
\Return{$\mathcal{P}_{\mathrm{coarse}}$}\;
\end{algorithm}

For each coarse seed $(\mathbf{d}_k^\star,\mathbf{c}_k^\star)$,
refinement fixes all discrete variables at $\mathbf{d}_k^\star$
and updates only the continuous sizing variables, initialized
at $\mathbf{c}_k^\star$.
Let $\hat{f}_i(\mathbf{d},\mathbf{c})$ denote the surrogate prediction
of objective $i$.
To account for their different units and magnitudes, we measure
objective changes relative to the corresponding coarse seed:
\begin{equation}
\begin{aligned}
\delta_i^{(k)}(\mathbf{c})
&=
\frac{
\hat{f}_i(\mathbf{d}_k^\star,\mathbf{c})
-
\hat{f}_i(\mathbf{d}_k^\star,\mathbf{c}_k^\star)
}{
\max\!\left\{
\left|\hat{f}_i(\mathbf{d}_k^\star,\mathbf{c}_k^\star)\right|,
\varepsilon
\right\}
},\\
s_i^{(k)}(\mathbf{c})
&=\lambda_i^{(k)}\delta_i^{(k)}(\mathbf{c}).
\end{aligned}
\label{eq:seed_residual}
\end{equation}
Here, $\varepsilon>0$ prevents division by zero, and
$\boldsymbol{\lambda}^{(k)}$ is a preference vector satisfying
$\lambda_i^{(k)}\geq 0$ and $\sum_{i=1}^{M}\lambda_i^{(k)}=1$.
Negative residuals indicate improvement over the coarse seed,
whereas positive residuals indicate deterioration.

At each refinement step, the residual scores define a softmax-weighted
gradient direction:
\begin{equation}
\mathbf{q}^{(k)}(\mathbf{c})
=
\sum_{i=1}^{M}
\frac{
\exp\!\left(s_i^{(k)}(\mathbf{c})/\mu\right)
}{
\sum_{j=1}^{M}
\exp\!\left(s_j^{(k)}(\mathbf{c})/\mu\right)
}
\nabla_{\mathbf{c}}s_i^{(k)}(\mathbf{c}),
\qquad \mu>0.
\label{eq:adaptive_gradient}
\end{equation}
Here, $\mu$ controls the smoothness of the gradient weighting.
Larger weighted residual scores receive larger coefficients,
adapting each objective's gradient contribution relative to the seed.

The normalized coefficients in
\autoref{eq:adaptive_gradient} form a softmax over the residual scores.
Using the softmax--log-sum-exp relation established
in~\cite{GaoPavel}, the aggregated gradient can be written as:
\begin{equation}
\mathbf{q}^{(k)}(\mathbf{c})
=
\nabla_{\mathbf{c}}\mathcal{L}_{\mu}^{(k)}(\mathbf{c}),
\qquad
\mathcal{L}_{\mu}^{(k)}(\mathbf{c})
=
\mu\log\!\left(
\sum_{i=1}^{M}
\exp\!\left(
\frac{s_i^{(k)}(\mathbf{c})}{\mu}
\right)
\right).
\label{eq:adaptive_potential}
\end{equation}

Compared with linear weighted scalarization, this form is capable of reaching unsupported solutions on non-convex regions of the Pareto front~\cite{STCH}.

\section{Experiments}
\label{sec:experiments}
Experiments run on an Intel Xeon Gold 6330 (2.00\,GHz) workstation with an NVIDIA A100 (40\,GB) GPU. The default technology is FreePDK45~\cite{FreePDK45} with $V_{\text{DD}}\!=\!1.0$\,V; process parameters ($V_{\text{th0}}$, $\mu_0$, $V_{\text{off}}$) use typical values with $\sigma\!=\!5\%$. SPICE runs use a parallelized Xyce engine~\cite{keiter2013xyce} at TT, \SI{25}{\degreeCelsius} unless noted, and all surrogate training data are generated from OpenYieldV2 to keep the evaluation chain self-contained. In the optimizer comparisons, all methods use the same TabPFN surrogate
conditioned on 500 SPICE samples, with 1{,}000 surrogate evaluations per run. 
Pareto-front quality is measured by hypervolume (HV) and inverted
generational distance (IGD).
Let $\mathcal{A}$ denote the obtained non-dominated objective vectors
in the evaluation space, with all objectives minimized.
Given an HV reference point $\mathbf{r}$ and an IGD reference set
$\mathcal{R}$, the indicators are defined as
\begin{equation}
\mathrm{HV}(\mathcal{A};\mathbf{r})
=
\operatorname{Vol}\!\left(
\bigcup_{\mathbf{a}\in\mathcal{A},\,\mathbf{a}\preceq\mathbf{r}}
\prod_{i=1}^{M}[a_i,r_i]
\right),
\label{eq:hv}
\end{equation}
\begin{equation}
\mathrm{IGD}(\mathcal{A};\mathcal{R})
=
\frac{1}{|\mathcal{R}|}
\sum_{\mathbf{z}\in\mathcal{R}}
\min_{\mathbf{a}\in\mathcal{A}}
\|\mathbf{z}-\mathbf{a}\|_2.
\label{eq:igd}
\end{equation}
Here, $\operatorname{Vol}$ denotes $M$-dimensional volume,
and $\preceq$ denotes componentwise inequality.
Higher HV and lower IGD indicate better front quality relative
to the chosen references.

\subsection{Compiler Verification}
\label{sec:exp_compiler}

\begin{figure}[tb]
\setlength{\abovecaptionskip}{0pt}
\centering
\includegraphics[width=1.0\linewidth]{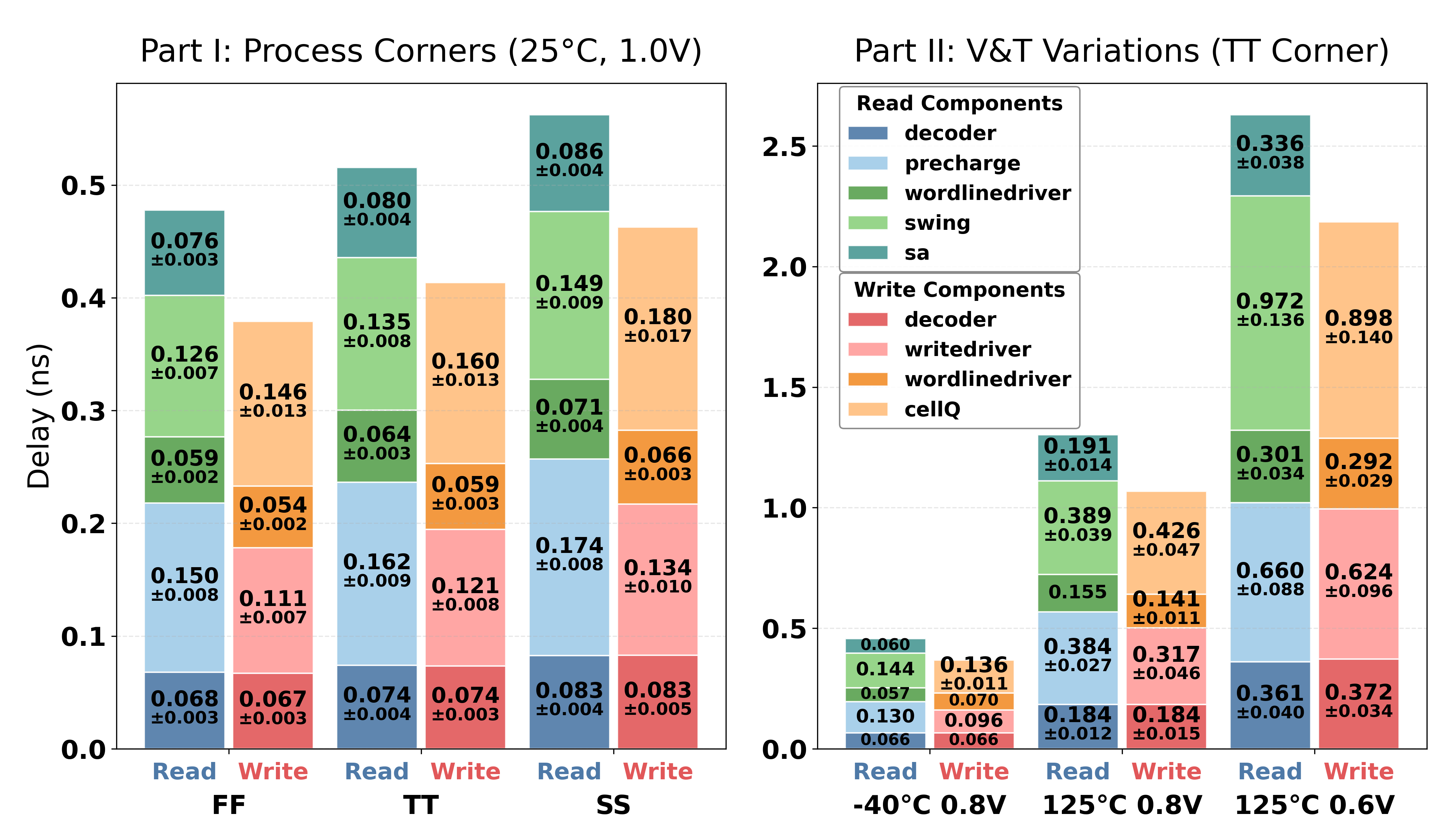}
\captionof{figure}{Mean and standard deviation of read/write delay across PVT corners for a 6T SRAM with parasitics, from 100 Monte Carlo simulations.}
\label{fig:Read&Write delay PVT}
\end{figure}

\autoref{tab:snm_delay_comparison_combined} compares hold/read/write SNM and timing for the 6T and 10T cells at nominal (\SI{1.0}{\volt}) and near-threshold (\SI{0.5}{\volt}) supplies under TT, \SI{25}{\degreeCelsius}. SNM is extracted by the standard butterfly-curve method~\cite{Kanj2009statistical, agarwal2008statistical}.
\begin{table}[b]
  \centering
  \caption{SNM and timing comparison for 6T and 10T cells at nominal and near-threshold supply voltages (TT, \SI{25}{\degreeCelsius}).}
  \label{tab:snm_delay_comparison_combined}
  \resizebox{0.8\columnwidth}{!}{%
  \begin{tabular}{clccccc}
    \toprule
    \multirow{2}{*}{\textbf{VDD}} & \multirow{2}{*}{\textbf{Metric}}
      & \multicolumn{2}{c}{\textbf{6T Cell}}
      & \multicolumn{2}{c}{\textbf{10T Cell}}
      & \multirow{2}{*}{\textbf{Unit}} \\
    \cmidrule(lr){3-4}\cmidrule(lr){5-6}
      & & \textbf{Mean} & \textbf{Std}
      & \textbf{Mean} & \textbf{Std} & \\
    \midrule
    \multirow{5}{*}{\rotatebox[origin=c]{90}{1.0\,V}}
    & Hold SNM    & 0.347 & 0.010 & 0.351 & 0.008 & V \\
    & Read SNM    & 0.152 & 0.013 & 0.171 & 0.013 & V \\
    & Write SNM   & 0.797 & 0.026 & 0.878 & 0.023 & V \\
    & Read $T_{\text{swing}}$  & 0.135 & 0.008 & 0.141 & 0.005 & ns \\
    & Write $T_{\text{cellQ}}$ & 0.160 & 0.013 & 0.162 & 0.011 & ns \\
    \midrule
    \multirow{5}{*}{\rotatebox[origin=c]{90}{0.5\,V}}
    & Hold SNM    & 0.187 & 0.007 & 0.230 & 0.004 & V \\
    & Read SNM    & 0.071 & 0.021 & 0.110 & 0.008 & V \\
    & Write SNM   & 0.229 & 0.022 & 0.250 & 0.017 & V \\
    & Read $T_{\text{swing}}$  & 0.546 & 0.072 & 0.795 & 0.060 & ns \\
    & Write $T_{\text{cellQ}}$ & 0.476 & 0.076 & 0.487 & 0.057 & ns \\
    \bottomrule
  \end{tabular}%
  }
\end{table}
At \SI{1.0}{\volt} the two topologies are comparable on hold stability (HSNM 0.347\,V versus 0.351\,V), but the gap widens sharply as the supply drops. At \SI{0.5}{\volt}, the 10T cell improves HSNM by 23.0\%, read SNM by 54.9\% (0.071\,V to 0.110\,V), and write SNM by 9.2\%, at the cost of a read delay that grows from 0.546\,ns to 0.795\,ns. \autoref{fig:Read&Write delay PVT} extends this comparison across PVT corners: read and write delay increase by roughly 22\% from the fastest (FF) to the slowest (SS) corner, and degrading from the best (\SI{-40}{\degreeCelsius}, 0.8V) to the worst (\SI{125}{\degreeCelsius}, 0.6V) operating condition pushes total read delay from 0.46\,ns to 2.63\,ns.

The two topologies, in other words, do not simply differ in overall quality; they trade advantage back and forth depending on supply voltage and corner. Neither is uniformly better, which is precisely the condition under which fixing one in advance discards information the optimizer could otherwise use. Section~\ref{sec:exp_boundary} asks whether this trade reversal is systematic enough to be characterized rather than left to intuition.

\subsection{Surrogate Model Evaluation}
\label{sec:exp_surrogate}

\begin{figure}[t]
    \centering
\includegraphics[width=0.88\columnwidth]
    {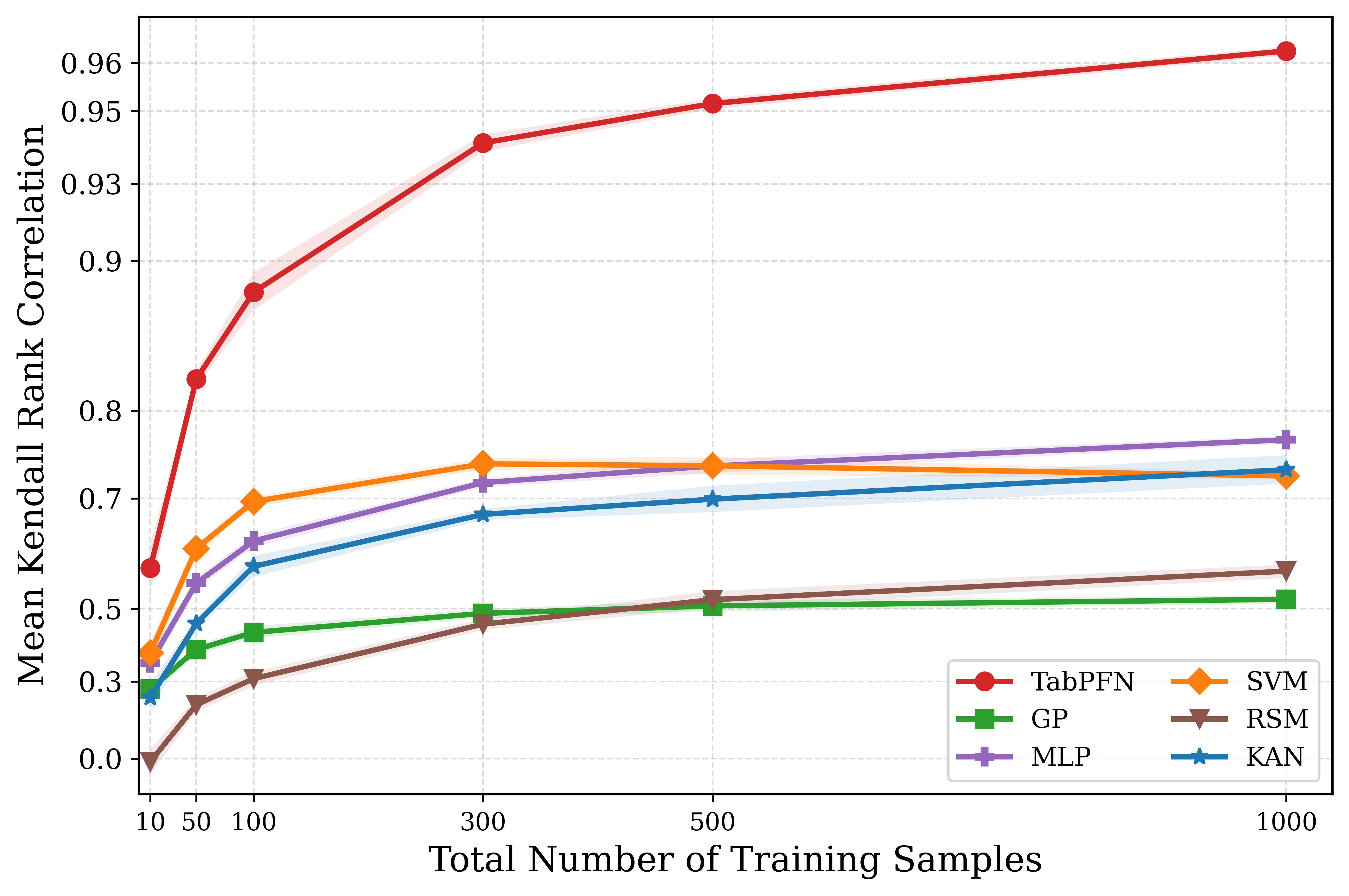}
    \caption{Kendall rank correlation ($\tau$) of different surrogates versus the number of training samples.}
    \label{fig:tabpfn_tau}
\end{figure}

\autoref{fig:tabpfn_tau} compares the ranking accuracy of six
surrogate models under different SPICE-sample budgets. TabPFN
reaches a mean Kendall correlation above 0.94 with 300 training
samples, whereas GP, MLP, SVM, KAN, and RSM all remain
below 0.80 even when the training budget is increased to 1{,}000
samples. 

We further evaluate whether the unified surrogate can reuse knowledge
across different topologies. Both 6T and 10T designs are represented
in the same input space with topology identity as a discrete feature.
For a data-limited target topology, the model starts with only 50
target samples and incorporates context from the other topology.
As shown in \autoref{tab:cross_topology_reuse}, reusing 6T context
reduces the 10T Macro-WAPE from $10.20\%$ to $5.73\%$ (43.9\%
reduction), while reusing 10T context reduces the 6T error from
$9.83\%$ to $4.48\%$ (54.5\% reduction). These results demonstrate
that TabPFN can effectively exploit shared information across SRAM
topologies, reducing the additional simulation data required for
new topology exploration.

\begin{table}[t]
    \centering
    \caption{Cross-topology knowledge reuse in a unified TabPFN surrogate.}
    \label{tab:cross_topology_reuse}

    \resizebox{\columnwidth}{!}{%
    \begin{tabular}{@{}lccccc@{}}
        \toprule
        \multirow{2}{*}{\textbf{Target}}
        & \multicolumn{4}{c}{\textbf{Macro-WAPE (\%) $\downarrow$(Target + Cross-Topology)}}
        & \multirow{2}{*}{\textbf{Reduction (\%) $\uparrow$}} \\
        \cmidrule(lr){2-5}
        & \textbf{50+0}
        & \textbf{50+100}
        & \textbf{50+200}
        & \textbf{50+500}
        & \\
        \midrule

        10T
        & $10.20 \pm 1.61$
        & $6.78 \pm 0.74$
        & $6.08 \pm 0.46$
        & $\mathbf{5.73 \pm 0.15}$
        & $\mathbf{43.9}$ \\

        6T
        & $9.83 \pm 1.37$
        & $6.58 \pm 0.88$
        & $5.37 \pm 0.27$
        & $\mathbf{4.48 \pm 0.37}$
        & $\mathbf{54.5}$ \\

        \bottomrule
    \end{tabular}%
    }
\end{table}


\subsection{The Topology Preference Boundary Analysis}
\label{sec:exp_boundary}

To understand the impact of exposing topology as an optimization
variable, we analyze how the optimizer selects between 6T and 10T
under different design conditions. Rather than fixing topology
beforehand, the joint framework reveals condition-dependent topology
preferences along the Pareto front.

\begin{figure}[t]
    \centering
    \includegraphics[width=0.75\linewidth]
    {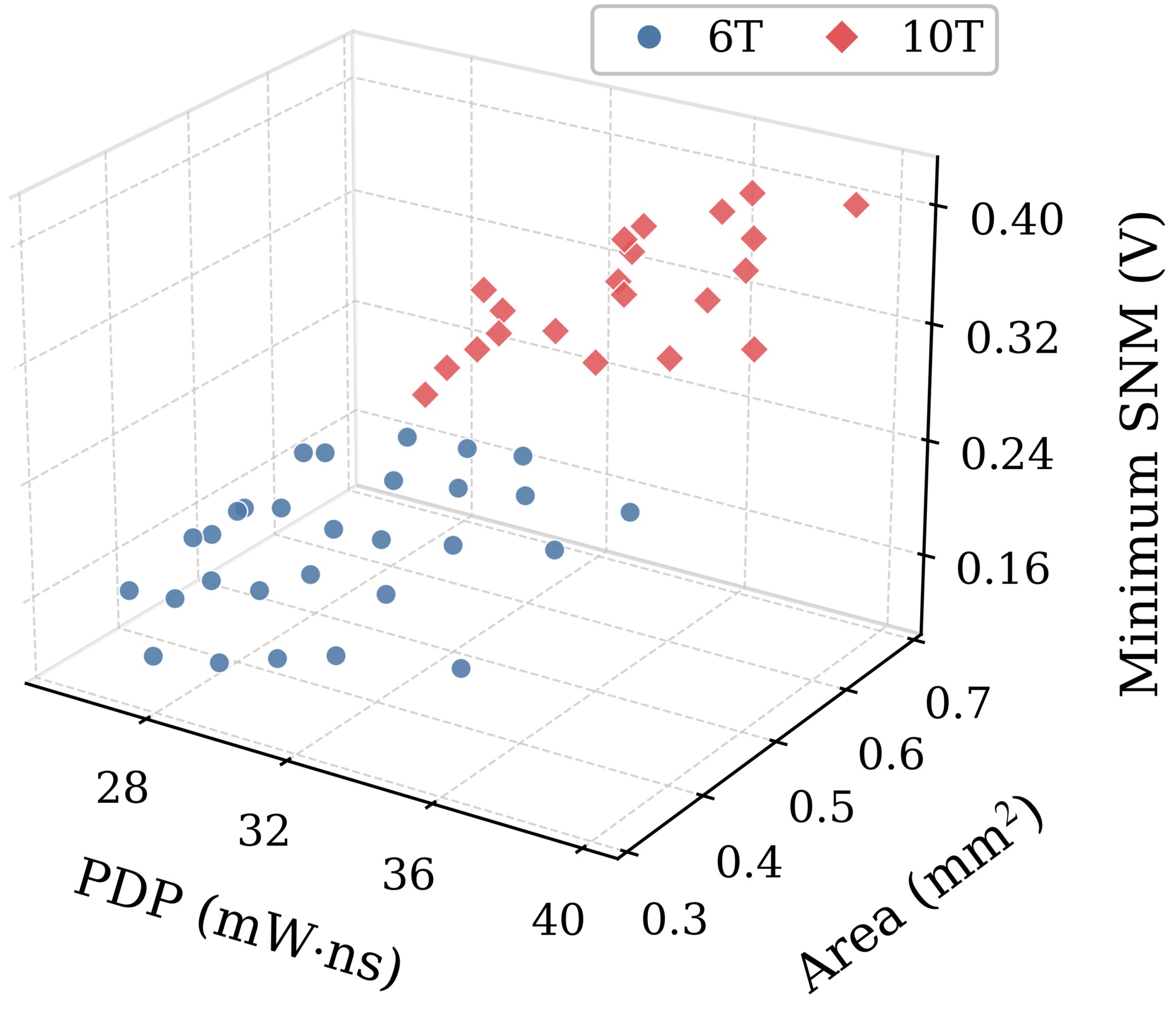}
    \caption{Joint search recovers both topologies on a single front.}
    \label{fig:joint_front_3d}
\end{figure}

\textbf{Joint search recovers both topologies on a single front.}
Running the proposed coarse-to-refined optimizer with topology and
threshold-voltage assignment as free variables produces the global
Pareto front shown in Figure~\ref{fig:joint_front_3d}. The coexistence
of both topologies indicates that neither uniformly dominates the other
across the complete design space. Instead, each topology contributes
solutions to different trade-off regions, which would be inaccessible
when the alternative topology is fixed.

\begin{figure}[t]
\centering
\begin{subfigure}[b]{0.495\linewidth}
\includegraphics[width=\linewidth]{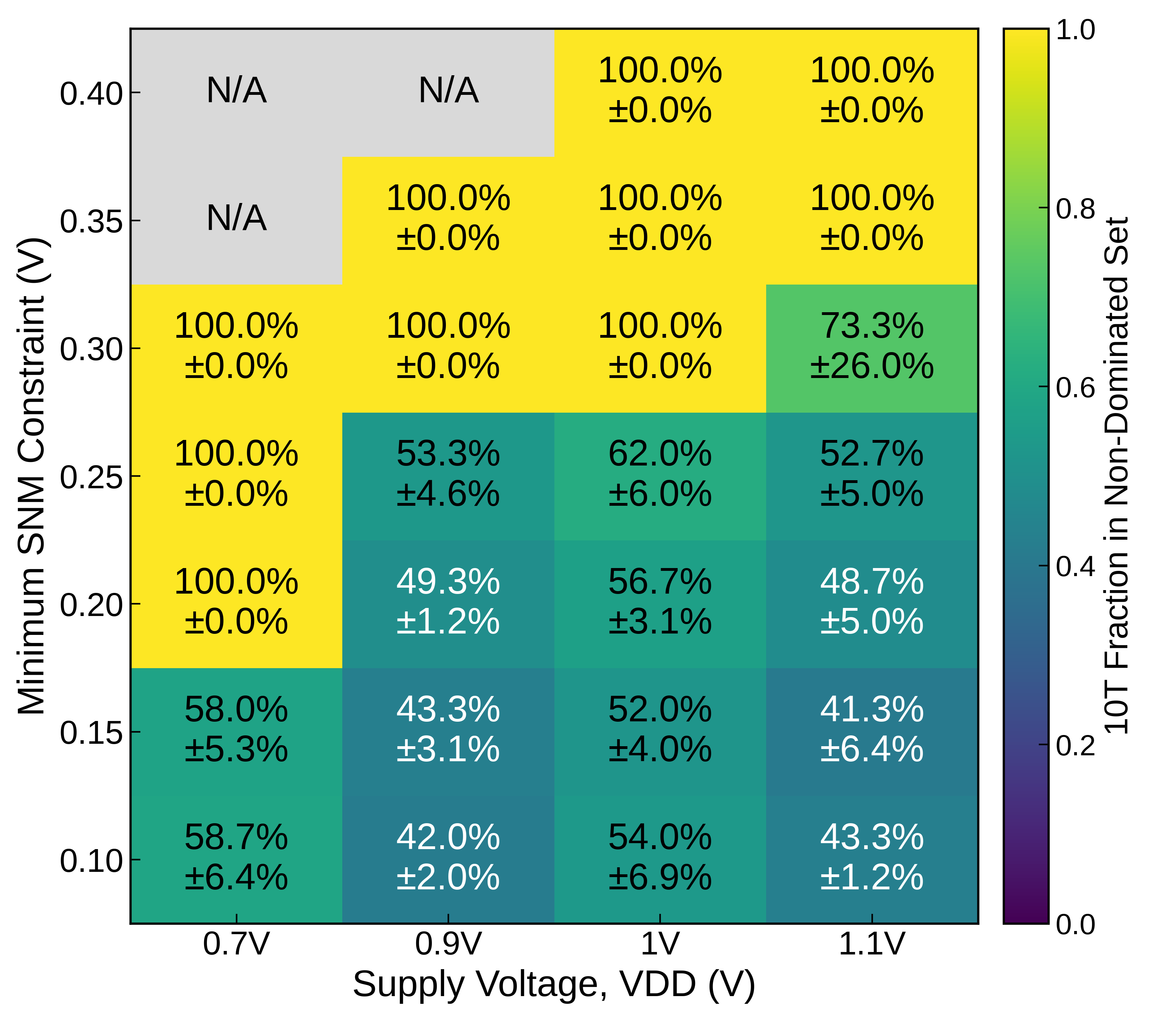}
\caption{Topology distribution under different VDD and SNM constraints.}
\label{fig:snm_vdd}
\end{subfigure}
\hfill
\begin{subfigure}[b]{0.495\linewidth}
\includegraphics[width=\linewidth]{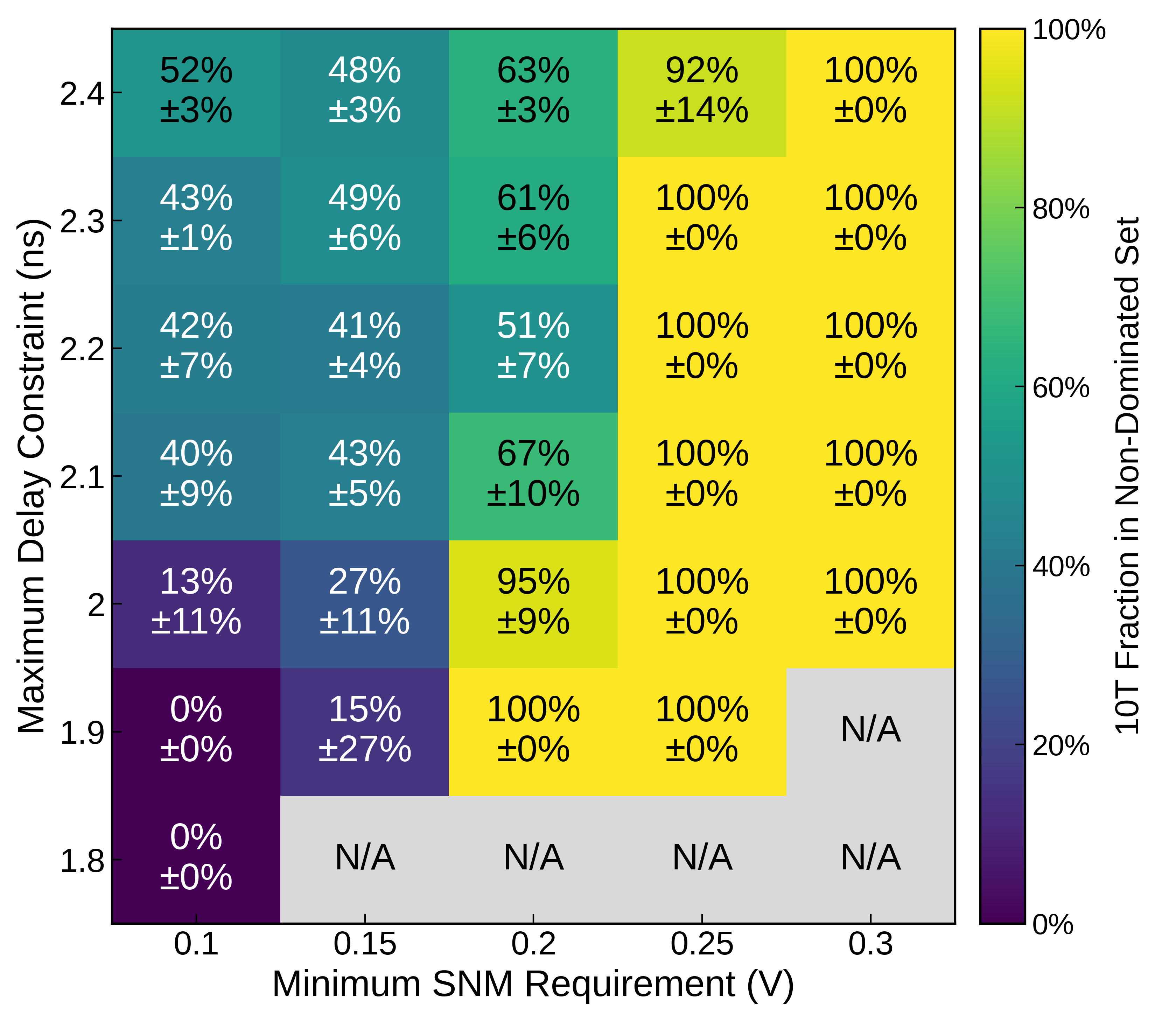}
\caption{Topology distribution under different SNM and delay constraints.}
\label{fig:snm_delay}
\end{subfigure}
\caption{
Topology decision boundaries between 6T and 10T SRAM under different design requirements.
The color represents the fraction of 10T solutions in the non-dominated set.
}
\label{fig:topology_boundary}
\end{figure}

\textbf{The optimal topology shifts with design conditions rather than following a fixed rule.}
Figure~\ref{fig:topology_boundary} reports the fraction of 10T solutions
in the non-dominated set while sweeping supply voltage and SNM-delay
constraints. Broad regions contain solutions from both topologies,
whereas topology-exclusive regions appear only under restrictive
requirements. The resulting boundary changes with design conditions
rather than following a deterministic rule. Therefore, a heuristic rule
derived under one condition may not remain reliable under another,
motivating topology to remain inside the optimization loop.

\textbf{Joint search outperforms heuristic topology selection and fixed-budget separate search.}
To directly test whether joint topology optimization is more effective than
separately optimizing each topology and merging the results, we compare four
methods under the same total budget: 6T-only search, 10T-only search,
Separate-Union with 500 evaluations assigned to each topology, and the
Joint search with 1{,}000 evaluations. For this comparison, topology
resampling is disabled in Joint search; hence, the observed 6T/10T allocation
is determined by selection and inheritance rather than random topology flips.

Table~\ref{tab:joint_compare} shows that Joint search improves both HV and IGD
over Separate-Union in the two mixed-topology scenarios S1 and S2.
The resulting allocations differ from the fixed 50/50 split, demonstrating
that a fixed per-topology budget can waste evaluations when the useful
topology balance is condition dependent. In S3, the feasible front
is strongly 10T-dominant. Although a topology oracle could reasonably favor
10T-only search in such a case, Joint search still exceeds the fair
Separate-Union baseline by both metrics by allocating most evaluations to 10T.

\begin{table}[t]
    \centering
    \caption{Joint topology--sizing search versus topology-fixed and
    equal-split separate-search baselines under representative design
    scenarios.}
    \label{tab:joint_compare}
    \scriptsize
    \setlength{\tabcolsep}{1.5pt}
    \renewcommand{\arraystretch}{1.05}
    \begin{tabular*}{\columnwidth}{
        @{\extracolsep{\fill}} c l c c c @{}
    }
        \toprule
        \textbf{Scen.} & \textbf{Method} &
        \textbf{HV $\uparrow$} & \textbf{IGD $\downarrow$} &
        \textbf{6T/10T (\%)} \\
        \midrule

        \multirow{4}{*}{S1}
        & 6T-only & 0.648$\pm$0.011 & 0.222$\pm$0.001 & 100/0 \\
        & 10T-only & 0.358$\pm$0.007 & 0.455$\pm$0.003 & 0/100 \\
        & Separate-Union (500+500)
        & 0.731$\pm$0.018 & 0.071$\pm$0.008 & 50/50 \\
        & \textbf{Joint search}
        & \textbf{0.739$\pm$0.023} & \textbf{0.067$\pm$0.017}
        & \textbf{60.2/39.8} \\
        \midrule

        \multirow{4}{*}{S2}
        & 6T-only & 0.485$\pm$0.039 & 0.276$\pm$0.026 & 100/0 \\
        & 10T-only & 0.366$\pm$0.022 & 0.416$\pm$0.013 & 0/100 \\
        & Separate-Union (500+500)
        & 0.624$\pm$0.019 & 0.081$\pm$0.009 & 50/50 \\
        & \textbf{Joint (proposed)}
        & \textbf{0.631$\pm$0.039} & \textbf{0.070$\pm$0.010}
        & \textbf{48.3/51.7} \\
        \midrule

        \multirow{4}{*}{S3}
        & 6T-only & 0.402$\pm$0.041 & 0.656$\pm$0.038 & 100/0 \\
        & 10T-only & 1.186$\pm$0.028 & 0.177$\pm$0.174 & 0/100 \\
        & Separate-Union (500+500)
        & 1.109$\pm$0.065 & 0.113$\pm$0.030 & 50/50 \\
        & \textbf{Joint (proposed)}
        & \textbf{1.208$\pm$0.036} & \textbf{0.069$\pm$0.023}
        & \textbf{10.2/89.8} \\
        \bottomrule
    \end{tabular*}

    \vspace{2pt}
    \begin{minipage}{\columnwidth}
        \scriptsize\raggedright
        \textit{S1}: $25^\circ$C, 1.1\,V; SNM $\geq 0.20$\,V;
        delay $\leq 4$\,ns; power $\leq 20$\,mW; area $\leq 0.7$\,mm$^2$.

        \textit{S2}: $25^\circ$C, 1.1\,V; SNM $\geq 0.25$\,V;
        delay $\leq 4$\,ns; power $\leq 20$\,mW; area $\leq 0.7$\,mm$^2$.

        \textit{S3}: $0^\circ$C, 1.0\,V; SNM $\geq 0.25$\,V;
        delay $\leq 3$\,ns; power $\leq 20$\,mW; area $\leq 0.8$\,mm$^2$.
        All methods use 1{,}000 surrogate evaluations. Values are mean
        $\pm$ standard deviation over three seeds. Metrics use the full
        feasible first-evaluation archive and a common pooled empirical
        reference front within each scenario.
    \end{minipage}
\end{table}

\subsection{Comprehensive Evaluation of the Proposed Optimizer}
Building on the topology-selection analysis above, we evaluate the
proposed coarse-to-refined optimizer against established multi-objective
optimization methods under both fixed-topology and joint-topology
settings. In the fixed-topology setting, the bitcell topology is fixed
to 10T, while the remaining architectural choices and transistor sizes
are optimized. In the joint-topology setting, the optimizer additionally
selects between 6T and 10T bitcells. All methods use the same TabPFN
surrogate conditioned on 500 SPICE samples. Only candidates with
a minimum SNM of at least $0.1\,\mathrm{V}$, a maximum delay of at most
$4.0\,\mathrm{ns}$, and a maximum power of at most $20\,\mathrm{mW}$
are retained. We compare against the evolutionary algorithms
NSGA-II~\cite{deb2002nsga2}, U-NSGA-III~\cite{seada2015unsga3},
SPEA2~\cite{zitzler2001spea2}, and C-TAEA~\cite{li2019ctaea},
as well as the Bayesian optimization baselines
ParEGO~\cite{knowles2006parego}, MACE~\cite{lyu2018mace},
and GPBO~\cite{snoek2012practical}.

\begin{figure}[t]
    \centering
    \includegraphics[width=0.8\linewidth]
    {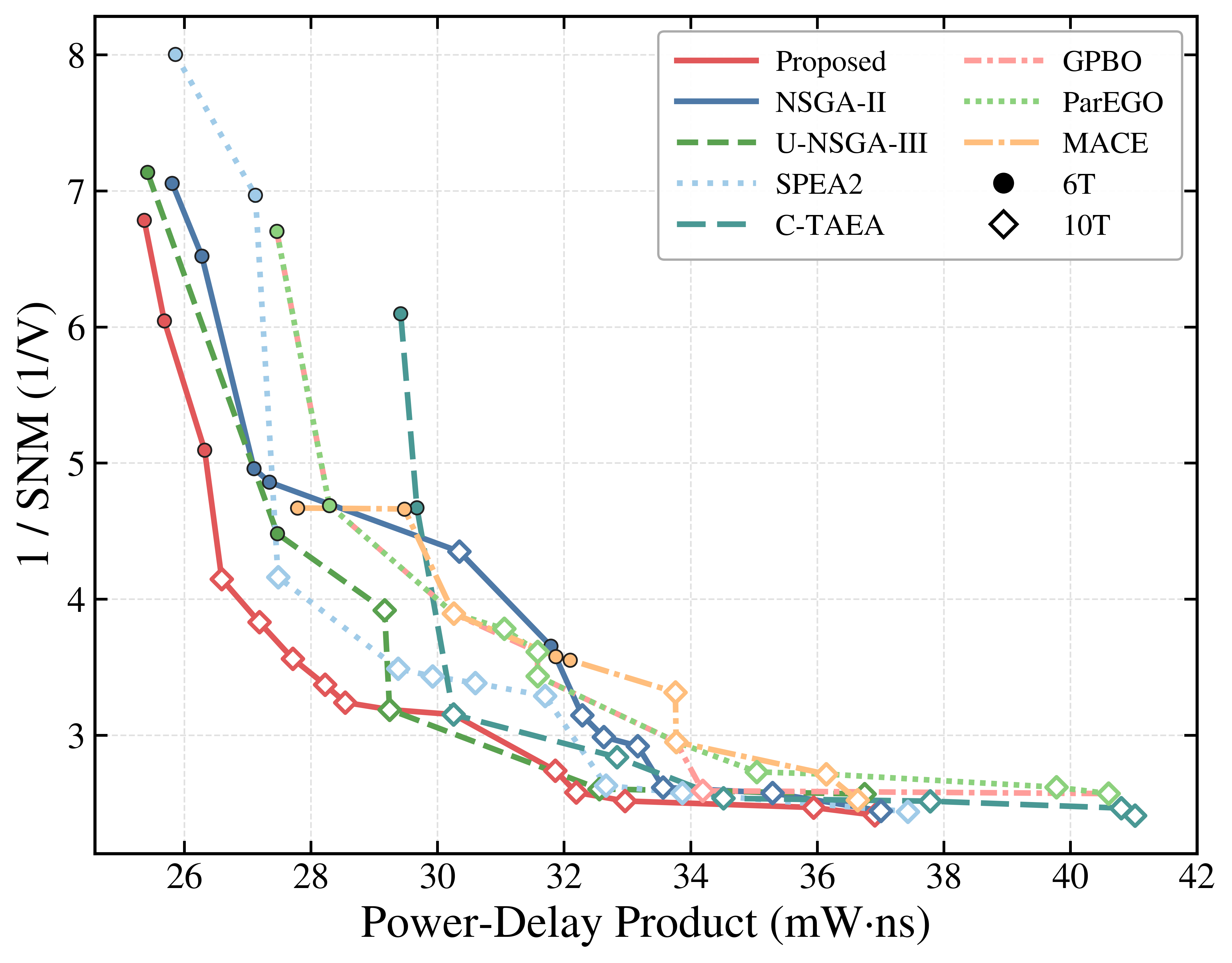}
    \caption{Joint topology–sizing optimization comparison with baseline optimizers.}
    \label{fig:optimizer_comparison_2d}
\end{figure}

Figure~\ref{fig:optimizer_comparison_2d} visualizes the
joint-topology search results in the PDP--inverse-SNM plane, while
Table~\ref{tab:optimization_comparison} reports HV and IGD for both
settings. With the topology fixed to 10T, the proposed optimizer
increases HV by 6.4\% and reduces IGD by 24.3\% relative to SPEA2,
the strongest baseline in this setting. These results show that
the proposed search strategy is effective even when topology is
selected in advance. Allowing topology to vary further improves
the results: the complete joint-topology approach increases HV
by 17.3\% and reduces IGD by 90.1\% relative to fixed-topology
SPEA2. Together, these comparisons support the benefits of
combining topology-aware exploration with continuous sizing
refinement.

Table~\ref{tab:optimization_comparison} further isolates the
contribution of gradient-based refinement by comparing the complete
optimizer with its coarse-search-only variant. Starting from the
same coarse-search results in the joint-topology setting, refinement
increases HV by 2.2\% and reduces IGD by 55.5\%. This paired
comparison indicates that local refinement provides additional
improvements to the surrogate-predicted Pareto front beyond those
obtained through coarse global exploration.

To complement the aggregate HV/IGD comparison with concrete circuit
instances, Table~\ref{tab:representative_pareto_designs} presents
the initial 6T reference design $I_0$ and seven representative
candidates from the joint search. Under nominal SPICE evaluation,
all seven candidates satisfy the SNM, delay, and power constraints.
Together with the model-estimated area, the results establish
mutual non-dominance among these candidates across the four
objectives. The three 6T and four 10T designs illustrate
complementary trade-offs obtained by varying bitcell topology,
transistor dimensions, threshold-voltage assignments, and
peripheral-circuit sizing.

\begin{table}[t]
    \centering
    \caption{Performance comparison of baseline algorithms and the proposed
    coarse-to-refine method under heuristic topology selection and joint
    topology optimization.}
    \label{tab:optimization_comparison} 
    \resizebox{0.8\linewidth}{!}{%
    \begin{tabular}{@{}lcccc@{}}
        \toprule
        \multirow{2}{*}{\textbf{Algorithm}}
        & \multicolumn{2}{c}{\textbf{Topology Fixed}}
        & \multicolumn{2}{c}{\textbf{Joint Optimization}} \\
        \cmidrule(lr){2-3}
        \cmidrule(lr){4-5}
        & \textbf{HV $\uparrow$}
        & \textbf{IGD $\downarrow$}
        & \textbf{HV $\uparrow$}
        & \textbf{IGD $\downarrow$} \\
        \midrule
        C-TAEA       & 0.818 & 0.2104 & 0.851 & 0.1093 \\
        ParEGO       & 0.780 & 0.2000 & 0.883 & 0.1058 \\
        MACE         & 0.774 & 0.1967 & 0.889 & 0.1302 \\
        GPBO         & 0.789 & 0.2207 & 0.893 & 0.1264 \\
        NSGA-II      & 0.790 & 0.1636 & 0.956 & 0.0672 \\
        U-NSGA-III   & 0.882 & 0.1588 & 0.993 & 0.0700 \\
        SPEA2        & 0.891 & 0.1438 & 0.952 & 0.0692 \\
        \midrule
        Proposed w/o refine & 0.946 & 0.1144 & 1.023 & 0.0319 \\
        \textbf{Proposed} & \textbf{0.948} & \textbf{0.1089}
        & \textbf{1.045} & \textbf{0.0142} \\
        \bottomrule
    \end{tabular}
    }
\end{table}

\begin{table*}[t]
\centering
\caption{Representative designs from the proposed joint 6T/10T topology--sizing search. $I_0$ is the initial 6T reference design, and $P_1$--$P_7$ are selected joint-search candidates that satisfy the electrical constraints under nominal SPICE evaluation. Transistor dimensions are in nm unless otherwise stated.}
\label{tab:representative_pareto_designs}
\resizebox{\textwidth}{!}{%
\begin{tabular}{@{}lcccccccc@{}}
\toprule
\textbf{Item} & \textbf{$I_0$} & \textbf{$P_1$} & \textbf{$P_2$} & \textbf{$P_3$} & \textbf{$P_4$} & \textbf{$P_5$} & \textbf{$P_6$} & \textbf{$P_7$} \\
\midrule
\multicolumn{9}{@{}c}{\textit{Architecture and bitcell variables}} \\
Topology & 6T & 6T & 6T & 6T & 10T & 10T & 10T & 10T \\
Array $R\!\times\!C\!\times\!N_{\mathrm{A}}$ & $128\!\times\!32\!\times\!64$ & $128\!\times\!128\!\times\!16$ & $128\!\times\!128\!\times\!16$ & $128\!\times\!128\!\times\!16$ & $128\!\times\!128\!\times\!16$ & $128\!\times\!128\!\times\!16$ & $128\!\times\!128\!\times\!16$ & $128\!\times\!128\!\times\!16$ \\
$W_{\mathrm{PU}}/W_{\mathrm{PD}}$ & 90.0/205.0 & 87.9/233.0 & 93.0/225.5 & 84.3/192.3 & 79.2/210.3 & 86.2/231.2 & 86.8/231.8 & 75.7/196.7 \\
$W_{\mathrm{PG}}/W_{\mathrm{FD}}$ & 135.2/-- & 127.5/-- & 115.8/-- & 129.6/-- & 118.0/152.4 & 148.9/161.2 & 145.6/160.5 & 110.1/161.1 \\
$L_{\mathrm{cell}}$ & 55.0 & 55.8 & 54.1 & 51.9 & 51.3 & 51.4 & 51.3 & 57.2 \\
$V_{\mathrm{T,PU}}/V_{\mathrm{T,PD}}/V_{\mathrm{T,PG}}/V_{\mathrm{T,FD}}$ & H/H/H/-- & G/G/G/-- & G/G/G/-- & G/H/H/-- & G/G/G/H & G/H/H/H & G/H/H/H & G/G/H/L \\
\midrule
\multicolumn{9}{@{}c}{\textit{Peripheral-circuit variables}} \\
Sense amplifier: $W_p/W_n/L$ & 540.6/269.8/55.0 & 455.6/297.6/53.0 & 441.6/301.8/51.2 & 441.9/274.0/50.1 & 444.7/308.6/51.5 & 441.5/318.5/50.5 & 444.8/315.1/50.7 & 452.4/305.0/50.2 \\
WL driver p: $W_{\mathrm{NAND}}/W_{\mathrm{INV}}/L$ & 269.6/269.7/55.0 & 315.6/280.9/51.6 & 320.1/252.3/50.6 & 320.3/236.0/50.2 & 319.1/259.2/50.4 & 320.3/228.1/50.6 & 318.2/220.7/50.8 & 221.4/288.7/51.0 \\
WL driver n: $W_{\mathrm{NAND}}/W_{\mathrm{INV}}$ & 179.7/89.8 & 204.7/74.6 & 211.9/73.5 & 213.1/74.2 & 214.1/99.0 & 214.8/76.0 & 214.6/76.9 & 215.0/87.6 \\
Precharge: $W_p/L$ & 270.6/55.0 & 266.5/56.2 & 279.2/53.7 & 234.5/56.3 & 322.8/55.6 & 219.8/50.4 & 222.0/51.0 & 220.2/59.7 \\
\midrule
\multicolumn{9}{@{}c}{\textit{SPICE-simulated electrical metrics and estimated area}} \\
Minimum SNM (V) & 0.250 & 0.209 & 0.215 & 0.274 & 0.248 & 0.342 & 0.343 & 0.421 \\
Maximum delay (ns) & 4.476 & 1.365 & 1.418 & 1.783 & 1.359 & 1.838 & 1.883 & 1.685 \\
Maximum power (mW) & 21.102 & 12.639 & 12.716 & 12.284 & 12.774 & 12.409 & 12.453 & 13.996 \\
Area (mm$^2$) & 0.4559 & 0.3629 & 0.3557 & 0.3428 & 0.5955 & 0.6224 & 0.6219 & 0.5949 \\
\bottomrule
\end{tabular}%
}
\vspace{1.5pt}
\parbox{0.98\textwidth}{\tiny\emph{Note:}
Electrical metrics are obtained using Xyce at the TT process corner,
25~$^\circ$C, and 1.0~V, with a capacity of 32~KB, 16 output
columns, and interconnect RC parasitics enabled. Area is estimated
using the analytical macro-geometry model.
$P_1$--$P_7$ satisfy minimum SNM $\geq 0.1$~V, maximum delay
$\leq 3.0$~ns, and maximum power $\leq 20$~mW, and are mutually
non-dominated across the four reported objectives.
$I_0$ is included only as an initial reference and is excluded from
this feasible set.
H/G/L denote high/general/low threshold-voltage models;
``--'' denotes a topology-inapplicable parameter.}
\end{table*}

\section{Conclusion}
\label{sec:conclusion}
We presented OpenYieldV2, which treats SRAM bitcell topology as a variable to be optimized jointly with transistor sizing rather than fixed in advance. A fully parameterized compiler generating both 6T and 10T macros under identical high-fidelity conditions, a few-shot TabPFN surrogate, and a coarse-to-refined optimizer together expose a topology decision boundary that shifts with design conditions rather than following a fixed threshold, and joint optimization across this boundary outperforms heuristic topology selection followed by separate sizing. This boundary is characterized here for 6T and 10T on FreePDK45; whether it holds a similar structure for additional bitcell topologies and advanced technology nodes remains open.

\bibliography{refs}
\bibliographystyle{ACM-Reference-Format}

\end{document}